\documentclass[runningheads]{llncs}
\usepackage{graphicx}

\usepackage{amsfonts}
\usepackage{amssymb}
\usepackage{amscd}
\usepackage{bbm}

\newcommand{\DB}{\mathit{D}}
\newcommand{\sel}{\mathit{sel}\xspace}

\newcommand{\sd}{\mathit{P}}

\usepackage{xspace}
\newcommand{\eg}{e.\,g.,\xspace}
\newcommand{\ie}{i.\,e.,\xspace}
\newcommand{\cf}{cf.\xspace}
\newcommand{\Eg}{E.\,g.,\xspace}

\usepackage{paralist}
\usepackage{cite}
\usepackage{moresize}

\begin{document}
\title{Local Exceptionality Detection in Time Series Using Subgroup Discovery}
\author{Dan Hudson\inst{1, 2}\orcidID{0000-0002-2917-4659}
\and\\
Travis J. Wiltshire\inst{2}\orcidID{0000-0001-7630-2695}
\and\\
Martin Atzmueller\inst{1}\orcidID{0000-0002-2480-6901}
}
\institute{Semantic Information Systems Group,
Osnabr\"uck University, Germany\\
\email{\{daniel.dominic.hudson, martin.atzmueller\}@uni-osnabrueck.de}\and
Department of Cognitive Science and Artificial Intelligence,\\
Tilburg University, The Netherlands \\
\email{t.j.wiltshire@tilburguniversity.edu}}

\maketitle
\begin{abstract}
In this paper, we present a novel approach for local exceptionality detection on time series data. This method provides the ability to discover interpretable patterns in the data, which can be used to understand and predict the progression of a time series. This being an exploratory approach, the results can be used to generate hypotheses about the relationships between the variables describing a specific process and its dynamics. We detail our approach in a concrete instantiation and exemplary implementation, specifically in the field of teamwork research. Using a real-world dataset of team interactions we include results from an example data analytics application of our proposed approach, showcase novel analysis options, and discuss possible implications of the results from the perspective of teamwork research.

\end{abstract}

\section{Introduction}

Methods for local exceptionality detection such as subgroup discovery~\cite{Atzmueller:15a} and its variant exceptional model mining (EMM)~\cite{Duivesteijn:2016:EMM:2877058.2877103} -- focusing on complex target models -- are established knowledge discovery techniques~\cite{FPS:96} for finding interpretable patterns. Basically, they identify patterns relating different attributes of a dataset that are interesting according to some target model; for example, these patterns could describe subgroups of a dataset having a high share of a given target variable. However, the application of such methods specifically to time series data has been limited in literature so far.

In this paper, we address this matter by presenting a novel approach for performing subgroup discovery and EMM on time series. We propose an extensible methodological approach and provide explanations and commentary on how to apply it, in particular relating to features and target construction on dynamic time series data, also introducing a novel quality function for subgroup discovery in the context of a respective data analytics application.

For demonstrating our approach, we exemplify its application on a case study conducted in the area of social sensing, wherein team interactions are examined through a multimodal, sensor-based approach. Body movements, including gesture and posture, along with dynamics of speaking and turn-taking, are well understood to be important social signals used in cooperation and teamwork \cite{wiltshire2019multiscale}. 
Relationships between these social signals could be useful for understanding how teams work together. However, it is difficult to establish this in an empirical way when using multiple time-varying modalities, based on recordings of teams. Using video and audio recordings, it is possible to extract signals that quantify body movement and speech volume over time, yet it is not obvious how to start analysing these time-varying signals to find meaningful relations within them. Therefore, the application of exploratory analysis methods such as subgroup discovery for mining interesting patterns is well suited for such an analysis, for first insights and to support hypothesis generation. 
Our approach leads to interpretable rules which are plausible due to the use of expert knowledge in feature selection (described further in Section \ref{subsec:feature-selection}). This is in contrast to, \eg neural network methods, which are able to learn predictive features from time series data, but are difficult to interpret into declarative rules and may learn features that are implausible as social signals, leading to poor hypothesis generation.  

We therefore choose this case study to showcase the ability of our method to generate hypotheses, investigating which combinations of features in body movement and speech data are potentially informative about teamwork dynamics. 
In summary, we present and interpret results from an analysis of 27 video and audio recordings of teams performing a collaborative task, taken from the ELEA corpus, \cf~\cite{sanchez2011nonverbal}. The discovered subgroups provide rules to relate the modalities of body movement and speech, contributing novel perspectives to the literature on teamwork with regard to how these modalities carry important information about interaction dynamics. This study demonstrates the usefulness of subgroup discovery with our novel analysis options as exploratory techniques.

\noindent Our contributions are summarised as follows:
\begin{enumerate}
\item We present a novel methodological approach as an iterative human-guided process that makes it possible to use subgroup discovery on time series data.
We discuss according feature extraction and target construction, %
\eg using different time lags, for making the results predictive at different timescales.
\item We showcase our approach through a study in the context of team research. For this analysis, we also introduce a new quality function, adapting the concept of dynamic complexity from team research. In addition, we present a novel subgroup visualisation for multi-dimensional parameter analysis, which we call the \emph{subgroup radar plot}, also enabling a user-guided \emph{Information Seeking Mantra}~\cite{Shneiderman:96} approach for subgroup assessment and comparison.
\item For our case study and results, we utilise a real-world dataset as an example data analytics application in which we search for relationships between multimodal data, \ie body movement and speech in time series.
As evaluated by a domain specialist, this gives rise to several meaningful hypotheses that can be investigated in future work in the field of teamwork study.
\end{enumerate}

The rest of the paper is organised as follows: Section~\ref{sec:related} summarises related work. After that, Section~\ref{sec:method} presents our method in detail. Next, Section~\ref{sec:case:study} introduces the applied case study and discusses our results. Finally, Section~\ref{sec:conclusions} concludes with a summary and outlines interesting directions for future work.

\section{Related Work}\label{sec:related}

Below, we first summarise related work on local exceptionality detection and subgroup discovery, before we sketch methods for time series analysis and finally introduce related approaches in our application context of teamwork analysis.

\subsection{Local Exceptionality Detection}

Methods for local exceptionality detection based on pattern mining typically aim to discover \emph{local models} characterising (or describing) parts of the data.
Then, not only an actionable model can be provided, but also a human-interpretable set of patterns~\cite{Mannila:00}.
Subgroup discovery~\cite{Kloesgen:96,Atzmueller:15a} is an approach for discovering interesting subgroups:
It aims at identifying subgroups of data instances that are \emph{interesting} with respect to a certain target concept, for example having a high chance of some interesting property being present. The interestingness of a subgroup is then accordingly defined by a certain property of interest formalised by a quality function.
Here, \emph{exceptional model mining}~\cite{Duivesteijn:2016:EMM:2877058.2877103} can be seen as a variant of subgroup discovery, focusing on more complex quality functions, \ie considering complex \emph{target models}, like comparing complex targets, regression models or graph structures on the levels of the respective local models.

A subgroup can be represented by a pattern which specifies membership in the subgroup, typically in the form of feature--value pairs, which must hold true for a data instance in order for it to be included in the subgroup. This means that it is a data-driven process that discovers explicit and \emph{interpretable} rules to associate descriptive properties found in the data instances, also enabling explanation-aware and further explainable data mining~\cite{AR:10a,Atzmueller:18:Declare}.
Feature--value pairs can be pre-determined via discretisation~\cite{Fayyad:1993gd}, or also be dynamically adapted from the data, \eg~\cite{meeng2021real}. We focus on the former for a user-guided process in which rules are simpler and therefore easier to navigate when using many features, making it easier for experts to propose subgroup rules and refine subgroup definitions. Moreover, this results in simpler rules which are easier to interpret and easier to generalise to other similar datasets, making them suitable hypotheses for further follow-on investigation. 

Considering time series and/or event data, the investigation of subgroup discovery has been limited, mainly focusing on aggregating/averaging time overall, \eg \cite{jin2014subgroup} or by considering aggregates on sets of discrete-valued events~\cite{knobbe2017sports}, compared to continuous time series which we consider in this work. For this, we present different techniques for feature as well as target construction embedded in a novel approach and process model.

\subsection{Exceptionality Detection in Time Series}

Time series exceptionality analysis is a vast field, \eg~\cite{aminikhanghahi2017survey,hamilton2020time}. Here, \eg methods for change detection~\cite{aminikhanghahi2017survey}, anomaly detection also including symbolic representations~\cite{keogh2005hot,chandola2010anomaly,RWA:19} and time series classification are relevant, where typically global approaches are addressed, in contrast to local exceptionality detection which we focus on in this work. For performing time series classification, a vast number of methods based on specific types of features have been proposed. An experimental evaluation of 37 different techniques is reported in~\cite{bagnall2017great}, grouped by the mechanism used for extracting the relevant features.
However, descriptive approaches using pattern mining on time series data have been mostly neglected so far, as also discussed above. An approach for compressing event logs based on the minimum description length (MDL) principle was presented by \cite{galbrun2018mining}, making it possible to detect local patterns in temporal data. Compared to this work, which focuses on event sequences, our approach aims to find meaningful representations of (potentially complex) continuous-valued recordings, and assesses the discovered patterns by reference to a target variable rather than compression. 

Common problems in time series (exceptionality) analysis involve feature construction~\cite{christ2016distributed} and feature selection~\cite{ATL13,GE2003}.
There exist standard approaches like \emph{TSFresh}~\cite{christ2018time} which generates a large set of possible features spanning from time-domain (\eg statistical measures such as mean and standard deviation) to using frequency-domain features constructed using Fourier transformations. This, however, can lead to the curse of dimensionality~\cite{ATL13} such that intelligent feature selection approaches are necessary, \eg~\cite{AHSK:17}.
In general, the idea is to select variables that are most informative;
common methods include ranking variables by correlation measures, one-class classifiers, or variable subset selection by wrappers or embedded methods~\cite{GE2003}. 
In contrast to such automatic approaches, we focus on interpretable local patterns, such that feature construction and selection are integrated into an iterative human-guided process. Then, according features can be selected, \eg based on domain knowledge or user feedback.

\subsection{Teamwork Analysis}

The modern workplace presents a complex socio-technical environment and often complex goals and objectives which must be tackled through collaborative,
effective teamwork.
The study of teamwork looks at how groups of multiple individuals work toward a common goal \cite{salas1992toward} through collaborative team processes \cite{fiore2010towards}. Recent work~\cite{kozlowski2018unpacking,salas2018science} has emphasised the need to understand dynamics within team processes, by embracing methodologies that record teams over time. In this vein, movement data has been analysed in the context of teamwork by \cite{wiltshire2019multiscale},
providing the sort of ongoing measurement of teams that are required to analyse time-varying dynamics and processes. Speech data of teams has been analysed, \eg by \cite{olguin2010assessing,kim2008meeting}
for better understanding team processes and performance. The case study we present shows one way that ongoing measurements of social signals (specifically body movement and speech recordings) can be analysed with a multimodal approach that can generate hypotheses about how to understand changes in team dynamics from the signals.

\section{Method}\label{sec:method}

Below, we first outline our approach, presenting a process model for local exceptionality detection in time series using subgroup discovery. After that, we discuss the individual steps in detail.

\subsection{Overview}

\begin{figure}
\includegraphics[width=1.0\textwidth]{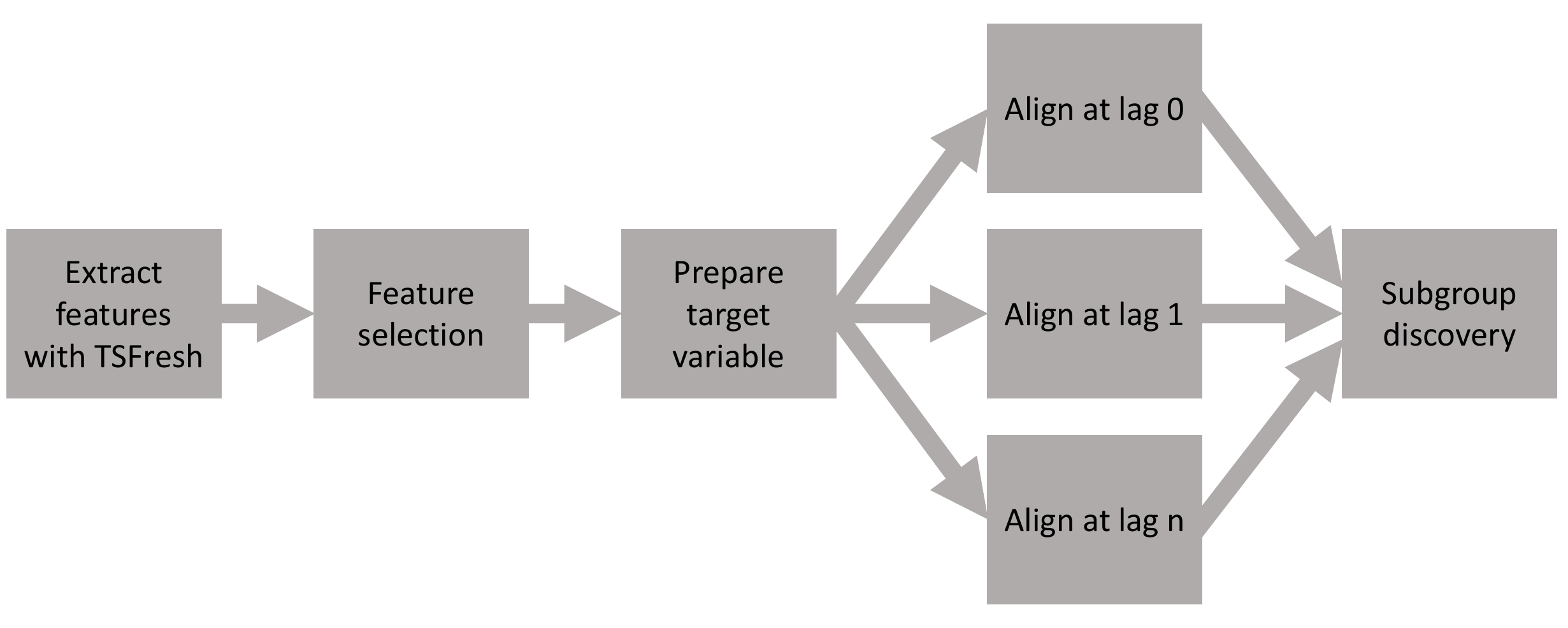}
\caption{The workflow of our methodological process to perform subgroup discovery on time series data. With a human-in-the-loop, the workflow can be iteratively applied.} \label{workflow}
\end{figure}

Our proposed approach is visualised in Figure \ref{workflow} as a linear workflow, which can be executed in multiple iterations: First, the time series is split into slices, i.e., non-overlapping subsequences of equal length, so that it is possible to investigate moments when a time-varying target variable reaches an extreme value.
For each slice, we extract a set of descriptive features.
These are (optionally) discretised, \eg using quantiles. \Eg the value of each feature can be converted into `low', `medium' and `high' based on tercile boundaries across the slices. The choice of the appropriate length for a slice should be driven by the application, \ie to include enough time points to allow a variety of features to be extracted, such as frequency components and estimates of entropy; also, it should be small enough that dynamics of the time series are not likely to change multiple times within a slice.
Then, the target variable is prepared. We also propose to examine different time lags between the features of time slices and the target, which necessitates performing the analysis with multiple copies of the dataset (with the lag applied).
In particular, what we propose is to investigate the relationship between attributes of the time slices to the target variable at different lags. With a lag of zero, our process discovers subgroups that are informative about how various attributes covary with the target. This could be useful when investigating, \eg how a system-level process is reflected in multiple variables. With higher lags, the process has a more predictive focus, asking which subgroups are particularly indicative of an especially high/low target value at a later point. 

\subsection{Time series feature extraction}

We convert time series into `slices' which can then individually be summarised with static rather than time-varying attributes.
To obtain the necessary features, we use the TSFresh package in Python \cite{christ2018time}, which computes a large number of features specifically to summarise time series.
Examples of features computed by TSFresh are, \eg:
\begin{inparaenum}[(a)]
    \item mean value,
    \item absolute energy,
    \item autocorrelation at different lags,
    \item Fourier coefficients,
    \item binned entropy,
    \item sample entropy,
    \item root mean square,
    \item etc.
\end{inparaenum}

This approach makes it possible to perform subgroup discovery on the features extracted for each slice, while still retaining some of the variation that is observed as the original time series progresses (since different slices will correspond to different points in time in the original series).

\subsection{Subgroup Discovery for Local Exceptionality Detection}

Subgroup discovery aims at finding a combination of criteria, in a form similar to rules (\eg \verb|PropertyA = True| or \verb|PropertyB > 1.5|), which then make up a pattern. Those rules are then used as membership criteria for a subgroup: any data points that satisfy the criteria are part of the subgroup. Specifying subgroups in this way is useful because the rules are easy to interpret and relate directly to known properties of the data points -- also called \textit{instances}.  
The key question is determining which subgroups are interesting using a quality function. Interesting subgroups could, \eg have a particularly high average target value compared to the population mean, as observed for the whole dataset.

More formally, \emph{selectors} or \emph{selection expressions} select the instances contained in a database $\DB$, which are covered by the respective patterns.
Typical selection expressions are given by attribute-value pairs in the case of nominal attributes, or by intervals in the case of numeric attributes.
A \emph{subgroup description} (or \emph{pattern}) combines selectors into a Boolean formula. For a typical conjunctive description language, a pattern $P = \{\sel_1, \ldots, \sel_{k}\}$ is defined by a set of selectors $\sel_j$, which are interpreted as a conjunction, i.e. $p =~\sel_1 \wedge \ldots \wedge~\sel_k$; each selector $\sel_k$ considers restrictions on the domain of an attribute contained in database $\DB$.
A \emph{subgroup} corresponding to a pattern then contains all instances $d \in \DB$ for which the respective formula for the pattern evaluates to true.
The interestingness of a pattern is determined by a quality function $q \colon 2^S \rightarrow \mathbb{R}\,.$ It maps every pattern in the search space to a real number that reflects the interestingness of a pattern (or the extension of the pattern, respectively).
Many quality functions for a single target feature, \eg in the binary or numerical case, trade off the size $n = |ext(\sd)|$ of a subgroup and the deviation $t_{\sd} - t_0$, where $t_{\sd}$
is the average value of a given target feature in the subgroup identified by the pattern $\sd$ and $t_0$ the average value of the target feature in the general population.
Thus, standard quality functions are of the form
\[
q_a (\sd)= n^a \cdot (t_{\sd} - t_0),\, a \in [0;1]\,.
\]
For binary target concepts, this includes, \eg a \emph{simplified binomial} function $q_{a}^{0.5}$ for $a=0.5$, or the \emph{Piatetsky-Shapiro} quality function $q_a^1$ with $a=1$, \cf~\cite{Atzmueller:15a}. Recently, \cite{boley2017identifying} described the use of quality functions which consider the dispersion of the target feature/variable within the subgroup, in addition to the average value of the target variable and the size of the subgroup. This refinement can be applied when it is desirable for subgroups to have consistency with respect to the target feature/variable.

\section{Results: Case Study on Team Interaction Data}\label{sec:case:study}

Below, we discuss a case study applying our approach in the context of interactive team cognition \cite{cooke2013interactive}. Here, our analysis is guided by the notion that interaction in teams changes, and should thus, be studied over time \cite{kozlowski2018unpacking, salas2018science}. We investigate the `dynamic complexity' of speech amongst team members (described in detail in Section \ref{target}), which is a method to quantify interaction dynamics which is also sensitive to moments of transition where the dynamics are changing. These moments are potentially to the benefit or the detriment of the team (we provide further discussion in \cite{wiltshire2020social}). Subgroup discovery provides interpretable patterns of interesting situations/events, where the dynamic complexity shows \emph{exceptional} local deviations, indicating interesting points in the respective team interaction.

\subsection{Dataset}

The data used in this case study comes from the Emergent Leadership (ELEA) corpus\cite{sanchez2011nonverbal}, which contains recordings of groups of individuals who have been tasked to work together to rank a list of items given a disaster scenario. In particular, the task was to rank the importance of items such as `chocolate' and `newspapers' for the situation in which the group has been stranded in a freezing-cold winter environment.
The corpus includes audio recordings from a single microphone in the centre of the room, and video recordings from webcams facing the participants. Both types of recording are available for 27 groups, each consisting of 3-4 participants. 
Via the video recordings, we quantify body movement during the team task.
See~\cite{hudson2021multisyncpy} for a detailed discussion on how to quantify body movement in this context.
Intuitively, we relate the body movement modality to the modality of speech, using the audio recordings to quantify speech dynamics. As a target we use dynamic complexity, \cf Section~\ref{target}, to estimate speech dynamics.

\subsection{Feature selection}\label{subsec:feature-selection}

TSFresh is not domain-specific, and therefore extracts generic features from time series. An important step in our process is to identify which features are potentially relevant and interpretable for the application being considered.
Here, it is worth considering whether the features succeed at capturing aspects of the time series that seem to relate to meaningful aspects of the process being studied.

\noindent We selected a subset of 91 from the 300+
features extracted by TSFresh. At a high level, these features can be categorised as follows: 

\begin{compactitem}
    \item Descriptive statistics (mean, variance, quantiles, standard deviations)
    \item Average and variance of the changes between successive time points
    \item Measures of complexity, such as `CID CE' and Lempel-Ziv complexity, as well as multiple forms of entropy
    \item `Matrix profile' statistics, which can be informative about repetitive or anomalous sub-sequences of the time slices
    \item Measures based on the number of peaks or extreme points
    \item Strength of different frequency components in the signal
    \item Measures based on autocorrelation at different lags
    \item Measures based on how well the data fits a certain (\eg linear) model
    
\end{compactitem}

Features were selected based on domain knowledge and in discussion with an expert in interactive team cogition: Selection criteria basically related to the potential relevance of the features to body movement in social interactions. For example, when considering coefficients of the Fourier transform, we included the magnitude since a large degree of movement at a specific frequency is something that can be visibly interpreted from the video recordings of body movement, but excluded the angle (or phase at the start of the time slice) since this is hard to visually comprehend (without performing further analysis to look for, e.g., synchrony) and therefore seems unlikely to be a usable social signal.

\subsection{Target Modeling -- Dynamic Complexity}\label{target}

As the target variable for subgroup discovery, we focused on the dynamic complexity of the speech recordings. The dynamic complexity measure is used to quantify how complex the behaviour of a system is, and provides us with a way to characterise the dynamics of speech, in a manner which could potentially be useful for (e.g.) detecting moments when a phase transition between patterns of behaviour is likely \cite{schiepek2010identification}.
Dynamic complexity is a combination of two components named the `Fluctuation' and the `Distribution', which are calculated over a (time) window of values. These quantities are originally presented in \cite{schiepek2010identification}, and we have described them and their use on speech data in \cite{wiltshire2020social}, which we follow below in our presentation and discussion of the basic concepts of dynamic complexity.

\textit{Fluctuation} ($F$) is based on finding `points of return' where the gradient stops being positive, zero, or negative (between these points we see increasing, unchanging, or decreasing subsequences of values), and then taking the difference in value of successive points of return. For each subsequence of increasing, unchanging or decreasing values, the absolute difference between the first and last elements is divided by the length of the sequence. This is summed across all relevant sub-sequences in the respective window. This sum is then normalised by dividing by a `maximum possible fluctuation' for the window size, which is the \textit{fluctuation} for a sequence that alternates between the maximum and minimum values. Frequent oscillations between high and low values lead to a higher value of $F$; consequently, a smaller-amplitude or less frequent oscillations lead to a smaller value.

\noindent Formally, the Fluctuation $F$ for a window with size $m$ is:

\begin{equation}
F = \left(\frac{1}{d(m - 1)}\right)
    \sum_{k=1}^{|P|-1} \frac{|p_{k+1} - p_{k}|}
                            {\phi(p_{k+1}) - \phi(p_k)}
\end{equation} 
where:

\begin{compactitem}
    \item $w = (x_1, \ldots, x_n)$ is a window containing a sequence of $n$ values $x_i \in X$ for a domain $X$
    with the corresponding index set $I = \{1, ..., n\}$
    \item $p = (x_j), j \in P, P \subseteq I$ is a subsequence of `points of return' including the start and end point of the window, using the definition of a `point of return' given above
    \item $\phi: X \times P \rightarrow I$ is a function which maps an element contained in the subsequence $P$ to its original position (index) in the window $w$. For example, if the $kth$ element $p_k$ contained in $P$ originally corresponds to the $jth$ element of $w$, i.e., $x_j$, then $\phi(p_k) = j$
    \item $d$ is the maximum possible difference between two values in $X$, i.e., $d = (x_{max} - x_{min})$, with $x_{max}$ and $x_{min}$ being the maximum and minimum values in the domain $X$.
\end{compactitem}

We compute the sum of absolute differences between subsequent elements $p$ in $P$ divided by the difference of their respective indices in $I$. This is then normalised according to window size $(m - 1)$ and maximal difference in values $d$. This quantity varies between 0 and 1.

\textit{Distribution} ($D$) is based on how equally-spaced the values in the window are after sorting. The sorted values of the window are compared to a sequence with maximum possible \textit{distribution} in which values are equally-spaced between the theoretical minimum and maximum. The disparity between these two sequences in terms of the spacing (e.g., comparing the difference between the first and third values within the first sorted sequence to the difference in the second, artificial, sequence) is found for pairs of indices, and then positive disparities are summed. This addition is performed across different sub-window sizes of the overall window. So, first, the differences at an offset of 1 (e.g., first and second sorted values) will be considered, followed by differences at an offset of 2 (e.g., first and third sorted values). Normalisation is achieved by dividing each term in the sum by the difference observed in the artificial sequence for that pair of indices. $D$ is highest when values are distributed equally across the measurement scale, and it decreases when there is a preference for certain ranges of values. Like fluctuation, this value varies between 0 and 1. 

The Distribution $D$ is then given as follows:

\begin{equation}
D = 1 - \sum_{c=1}^{m-1}\sum_{d=c+1}^{m}\sum_{a=c}^{d-1}\sum_{b=a+1}^{d} \frac{\Delta_{ba}\Theta(\Delta_{ba})}{\delta_{Y_{ba}}} \,,
\end{equation}
where:
\begin{itemize}
    \item $m$ is the length of the sequence
    \item $\Delta_{ba}$ is the difference between the values at indices $a$ and $b$ within the sorted values of the window
    \item $\Theta$ is the Heaviside step function: $1$ if the input is positive, $0$ otherwise
    \item $\delta_{Y_{ba}}$ is the difference between values at indices $a$, $b$ within the artificial values
\end{itemize}

\noindent The product of $F$ and $D$ gives the \textit{dynamic complexity} within the window. 

The dynamic complexity calculation takes a univariate time series as an input. Since the sampling rate of our audio data is particularly high, being 40,000Hz, we produce a more coarse-grained dataset by computing the energy of each second of audio. 
This allows us to calculate dynamic complexity using a reasonable window size while examining the evolution of the speech dynamics on a scale appropriate to interaction behaviour (a timescale of seconds to minutes). 

We evaluated two variations of dynamic complexity, captured as a target attribute, which we constructed in the spirit of EMM.
First, we model the dynamic complexity observed in each slice as a Gaussian distribution of values, and use the z-score normalised mean as the key parameter that will determine which subgroups are most interesting, \ie as our target for subgroup discovery. Second, we perform a linear regression of the dynamic complexity against time as a target model, and use the resulting slope as the target attribute. In both cases, the quality function we use to rank subgroups is the simple binomial quality function ($q_a^{0.5}$), which tends to favour smaller subgroups with a more extreme target value. To balance this, we experimented with different thresholds for a minimum subgroup size.
From our empirical analysis, we set the minimum subgroup size to 20 so that they do not become too small to be meaningful.

\subsection{Results}

\begin{figure}[ht]
\centering
\includegraphics[width=1.0\textwidth]{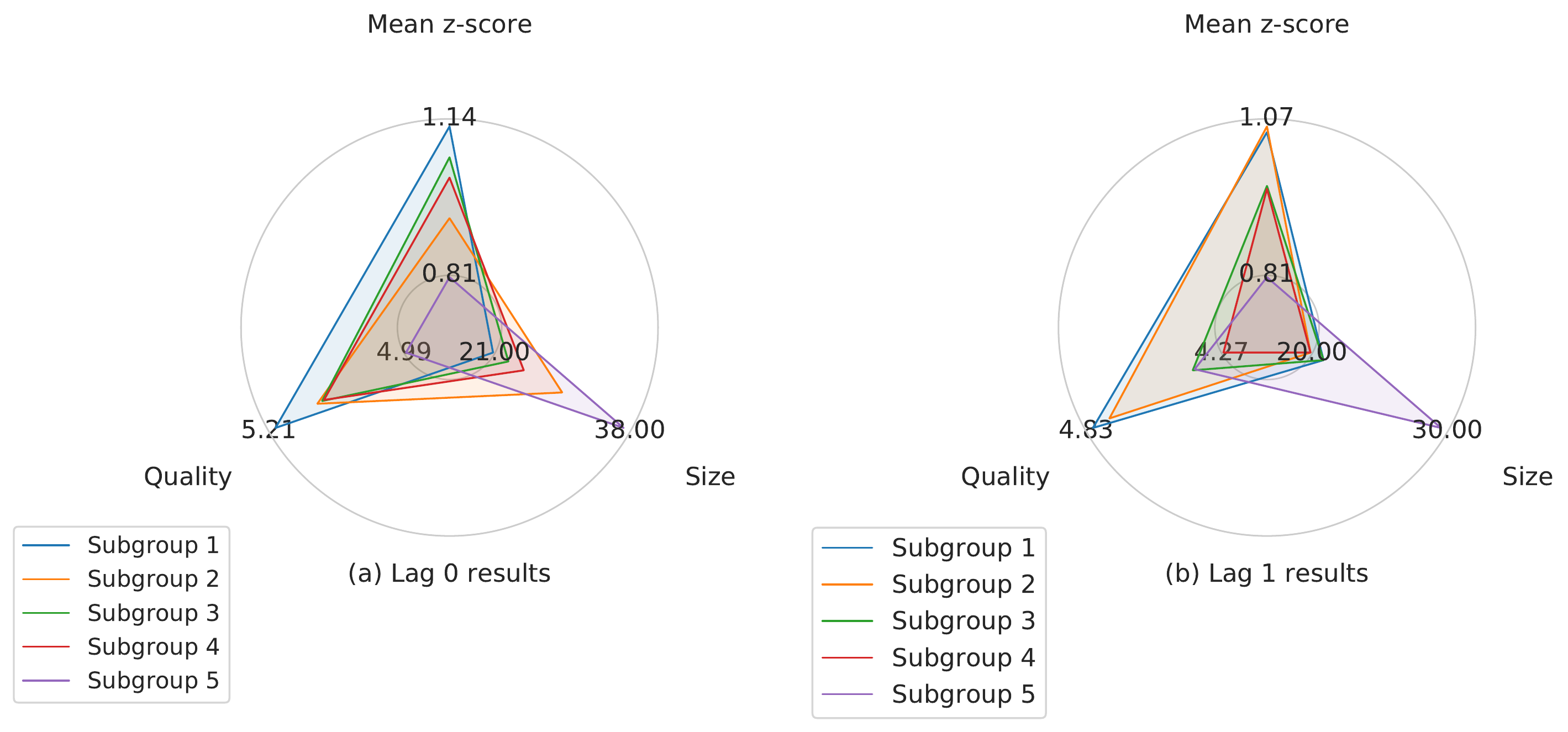}
\caption{Visualisations of how subgroups differ according to quality, mean z-score, and size. The results are shown at: (a) a lag of 0 minutes, and (b) a lag of 1 minute.} \label{fig:combined-radar-plots-quality}
\end{figure}

\begin{figure}[ht]
\centering
\includegraphics[width=1.0\textwidth]{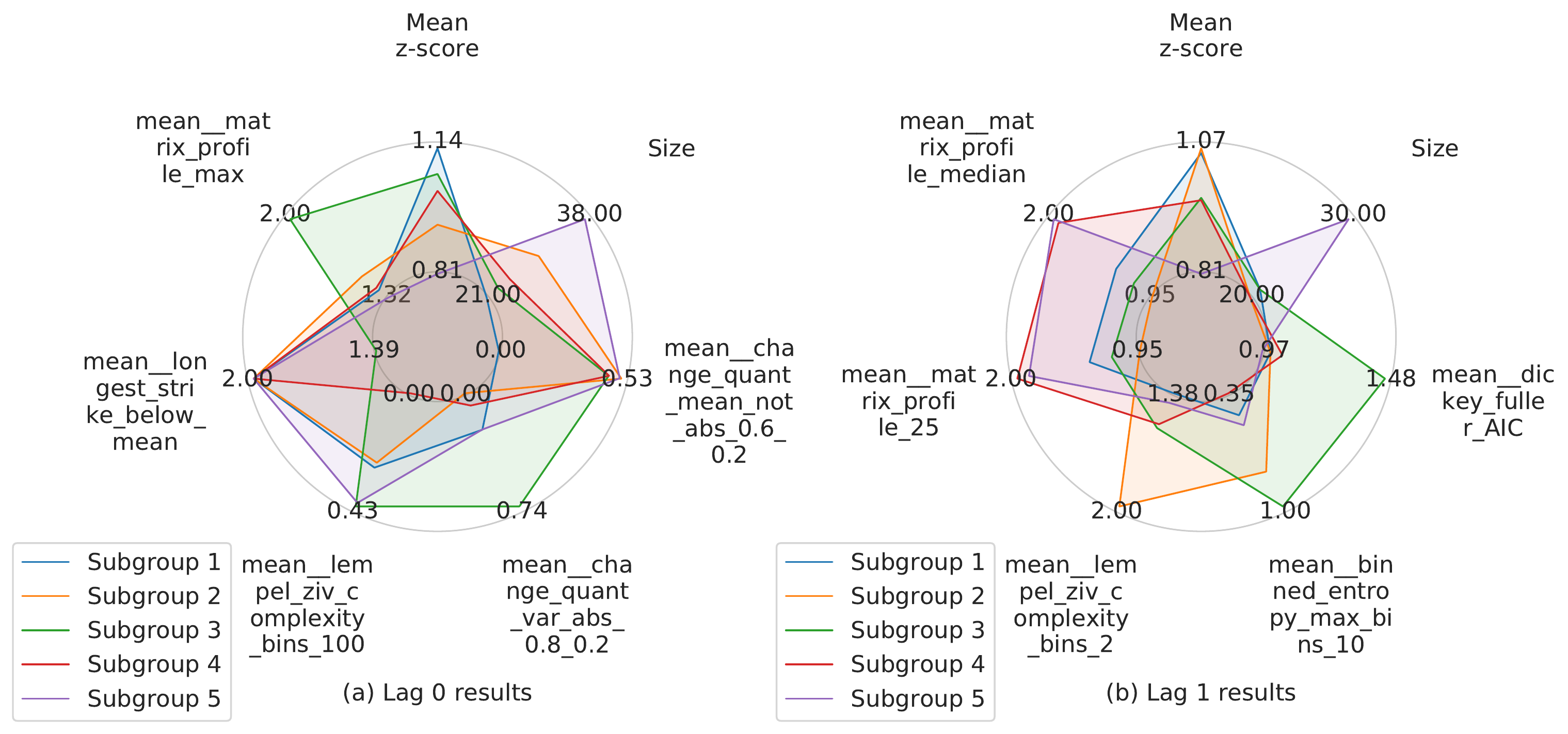}
\caption{Visualisations of how subgroups differ according to the mean z-score, the size, and 5 key selector variables -- with (a) a lag of 0 minutes, and (b) a lag of 1 minute.} \label{fig:combined-radar-plots}
\end{figure}

As stated earlier,
we are able to perform subgroup discovery at different time lags, making the task predictive when using lags greater than zero, or an exploration of the relationships between variables at lag zero. First, we discuss the 0-lag results, with slices of 1 minute. A selection of five subgroups is presented in Table~\ref{lag-0-table}; the subgroups are also visualised as a subgroup radar plot in Figure \ref{fig:combined-radar-plots-quality}(a),
indicating the most important quality parameters. Furthermore, Figure~\ref{fig:combined-radar-plots}(a) shows a detailed view on how subgroups differ according to the mean z-score, the subgroup size, and 5 key selector variables, as a ``zoom-in'' on the different parameter sets. Here, both quality and descriptor variables are combined, whereas both can also be inspected separately, in a human-in-the-loop approach.
This allows the user-guided inspection of a set of quality parameters (and/or subgroup selectors), for subgroup assessment and comparison.

In general, these presented novel subgroup visualisations (Figures~\ref{fig:combined-radar-plots-quality}-\ref{fig:combined-radar-plots}) allow a seamless overview--zoom--detail cycle, according to the  \emph{Information Seeking Mantra} by Shneiderman~\cite{Shneiderman:96}: Overview first (macroscopic view), browsing and zooming (mesoscopic analysis), and details on demand (microscopic focus) -- from basic quality parameters to subgroup description and its combination. This proved beneficial for our domain specialist when inspecting the results of our knowledge discovery methodology implemented via the iterative workflow.

Overall, our results show that it is possible to discover relatively small subgroups (size 20-40 compared to a population size of 327) whose dynamic complexity has on average a z-score of around 1, as the mean, averaged accross members of the subgroup.

\begin{table}
    \begin{ssmall}
    \caption{A selection of subgroups discovered using the SD-Map algorithm~\cite{AP:06} at a lag of 0 minutes: subgroup pattern, a textual description, size ($S$) and mean z-score ($\varnothing$).}\label{lag-0-table}    \begin{tabular}{|p{0.025\linewidth}|p{0.36\linewidth}|p{0.47\linewidth}|p{0.045\linewidth}|p{0.055\linewidth}|}
    \hline
     & Pattern & Description & $|S|$ & $\varnothing$ \\
    \hline
    \hline
    
    1 &
    \verb|mean__change_quantiles__f_agg_"mean"_| \verb|_isabs_False__qh_0.6__ql_0.2=low|, AND 
    \verb|mean__longest_strike_below_mean=high|, AND 
    \verb|mean__quantile__q_0.8=low| & The value of changes around the mean (after values have been restricted to remain between the 0.2 and 0.6 quantiles) is low across the team.  Team members tend to have at least one long sequence of values below the mean. The 0.8 quantile also tends to be low.  & 21 & 1.137 \\
    \hline
    
    2 & 
    \verb|mean__change_quantiles__f_agg_"var"_| \verb|_isabs_True__qh_0.8__ql_0.2=low|, AND 
    \verb|mean__longest_strike_below_mean=high|, AND 
    \verb|mean__quantile__q_0.8=low| & The absolute value of changes around the mean (after values have been restricted to remain between the 0.2 and 0.6 quantiles) is low across the team.  Team members tend to have at least one long sequence of values below the mean. The 0.8 quantile also tends to be low.  & 30 & 0.938 \\
    \hline
    
    3 &
    \verb|mean__matrix_profile__feature_"max"_|
    \verb|_threshold_0.98=high|, AND 
    
    \verb|std__lempel_ziv_complexity_| \verb|_bins_100=medium|, AND 
    
    \verb|std__quantile__q_0.3=low| & The maximum similarity of subwindows within the signal to other subwindows within the signal is low, suggesting that subsequences (of the time series) are all to some extent unusual (not repeating). The Lempel-Ziv measure of complexity is neither high nor low across the team. The 0.3 quantile does not vary much across the team members.  & 23 & 1.07 \\
    \hline
    
    4 &
    \verb|mean__lempel_ziv_complexity_| \verb|_bins_100=low|,  AND 
    
    \verb|mean__longest_strike_below_mean=high|, AND  \verb|mean__quantile__q_0.8=low| & The Lempel-Ziv measure of complexity is neither high nor low across the team. Team members tend to have at least one long sequence of values below the mean. The 0.8 quantile also tends to be low.  & 25 & 1.026 \\
    \hline
    
    5 &
    \verb|mean__longest_strike_below_mean=high|, AND  \verb|mean__mean=low| &  Team members tend to have at least one long sequence of values below the mean. The average value of movement is generally low across the team. & 38 & 0.81 \\
    \hline
    \end{tabular}
    
    \end{ssmall}
\end{table}

Since the outputs are interpretable, it is possible to speculate about what they mean in the context of body movement and speech dynamics. Many of the subgroups, such as those shown in Table \ref{lag-0-table}, suggest that low amounts of movement might be indicative of complex speech dynamics, particularly if there is a long sequence of low values. This perhaps suggests that while speech dynamics are becoming chaotic, the team members become more still -- for example moving less as they focus more on the discussion. This is a hypothesis generated from the data which further work could seek to verify. 
In order to consider the impact of the window size, we also performed subgroup discovery using 30-second slices of the time series. This uncovered subgroups which often used the same rules, e.g., stating that the change around the mean between the 0.2 and 0.6 quantiles be low across the team, and, that the mean and various quantiles should also be low. There were some differences, especially that these subgroups incorporated more rules concerning variability between team members with respect to their number of values below the mean/above the mean and their Lempel-Ziv complexity, suggesting that certain types of imbalance in the teams may also help to identify moments of high dynamic complexity in speech. Overall, changing the window size in this way did not have a large impact on the discovered subgroups.

Next, we discuss results when applying a lag of 1 minute, discovering ways to predict high complexity in speech dynamics from the body movement signals a minute earlier. A selection of subgroups are presented in Table \ref{lag-1-table},
Figure \ref{fig:combined-radar-plots-quality}(b), and Figure~\ref{fig:combined-radar-plots}(b), to give an idea of how exceptional the subgroups are compared to the population overall. Like with the 0-lag results, it appears to be possible to discover subgroups of around 20-40 members which have an average z-score of close to 1. The features used to define subgroups, however, are different. 

\begin{table}[ht]
    \begin{ssmall}
    \caption{A selection of subgroups discovered using the SD-Map algorithm~\cite{AP:06} at a lag of 1 minute: subgroup pattern, a textual description, size ($S$) and mean z-score ($\varnothing$).}\label{lag-1-table}
    \begin{tabular}{|p{0.025\linewidth}|p{0.36\linewidth}|p{0.47\linewidth}|p{0.045\linewidth}|p{0.055\linewidth}|}
    \hline
     & Pattern & Description & $|S|$ & $\varnothing$ \\
    \hline
    \hline
    
    1 &
    \verb|mean__augmented_dickey_fuller_| \verb|_attr_"teststat"__autolag_"AIC"=medium|, AND \verb|mean__spkt_welch_density__coeff_2| 
    \verb|=low|, AND 
    \verb|std__0.0_to_0.25_Hz=low| & The signal is neither relatively well-modelled nor relatively poorly-modelled by a process with a unit root, according to the Augmented Dickey-Fuller test. The strength of the frequency component at 0.234 Hz is low, and there is low variability of the strength of frequency components between 0.0 Hz and 0.25 Hz among team members.  & 21 & 1.062 \\
    \hline
    
    2 &
    \verb|mean__augmented_dickey_fuller_|  \verb|_attr_"teststat"__autolag_"AIC"=medium|, AND 
    \verb|mean__lempel_ziv_complexity_| 
    
    \verb|_bins_2=high|,  AND 
    
    \verb|mean__spkt_welch_density__coeff_2=low| & The signal is neither relatively well-modelled nor relatively poorly-modelled by a process with a unit root, according to the Augmented Dickey-Fuller test. The strength of the frequency component at 0.234 Hz is low, and there is low variability of the strength of frequency components between 0.0 Hz and 0.25 Hz among team members.  Complexity, according to a simple Lempel Ziv complexity score involving two bins, is high across the team. & 20 & 1.072 \\
    \hline
    
    3 &
    \verb|mean__binned_entropy_| 
    
    \verb|_max_bins_10=medium|, AND  
    
    \verb|mean__ratio_beyond_r_sigma__r_1.5=high|, AND 
    \verb|mean__skewness=medium| & Complexity according to binned entropy is neither high nor low across the team. The proportion of values more than 1.5 sigma from the mean is high. The distributions of values are generally neither highly nor lowly skewed.  & 21 & 0.969 \\
    \hline
    
    4 &
    \verb|mean__matrix_profile__feature_"25"_| \verb|_threshold_0.98=high|, AND 
    
    \verb|mean__spkt_welch_density__coeff_2=low|, AND 
    \verb|std__root_mean_square=low| & The 0.25 quantile of the similarity of subwindows within the signal to other subwindows within the signal is low, suggesting that a reasonable proportion of subsequences (of the time series) are unusual (not repeating). The strength of the frequency component at 0.234 Hz is low. How well the time slices can be modelled by a linear progression is not varied across the team members.  & 20 & 0.964 \\
    \hline
    
    5 &
    \verb|mean__matrix_profile__feature_"median"_| \verb|_threshold_0.98=high|, AND 
    
    \verb|std__change_quantiles__f_agg_"var"_| \verb|_isabs_True__qh_0.6__ql_0.4=low|, AND 
    
    \verb|std__quantile__q_0.9=low| & The median similarity of subwindows within the signal to other subwindows within the signal is low, suggesting that subsequences (of the time series) tend to be relatively unusual (not repeating). The absolute value of changes around the mean (after values have been restricted to remain between the 0.4 and 0.6 quantiles) is consistent across the team. The 0.9 quantile also has low variability.  & 30 & 0.811 \\

    \hline
    \end{tabular}
    
    \end{ssmall}
\end{table}

Some of the subgroups, such as the first two listed in Table \ref{lag-1-table}, suggest that the team members might have similar, low values for the low-frequency movement components during the minute preceding a period of high dynamic complexity. Features based on `matrix profile' statistics are used to define many subgroups. Looking at these features, it seems that high dynamic complexity can be expected following a period when the body movement signal does not have clear repetitions in its structure. This could indicate that complexity and a lack of pattern in body movement is predictive of chaotic speech dynamics shortly thereafter. This is another example of a data-driven hypothesis that future confirmatory work could verify.

We also re-ran this analysis using slices with a duration of 30 seconds, to detect any differences caused by window length. These subgroups also indicated that a low strength at 0.234 Hz could help to predict high complexity in speech dynamics, but they additionally include rules where there is a high frequency component at 0.585 Hz. Matrix profile features were often medium in these subgroups, suggesting that with smaller window lengths, a lack of repetition is not so clearly found. Other rules, for example that the number of values beyond 1.5 standard deviations from the mean should be high, were consistent across the 30-second and 1-minute window sizes.

Furthermore, we performed the analysis with different target concepts besides the mean dynamic complexity of speech. First, we considered the slope when conducting a linear regression of the dynamic complexity against time. The results in this case would be informative about periods when the complexity of speech dynamics is increasing. Many of these subgroups included rules that complexity of body movement (measured through Lempel-Ziv, Fourier entropy and binned entropy) is generally high across the team, and some subgroups also indicated that a large number of peaks in the slices of body movement was related to increasing dynamic complexity in speech. This could indicate that complexity in speech increases fastest while movement is already complex. From this analysis, the subgroups also suggested a relationship between increasing speech complexity and a medium-strength frequency component in body movement at 0.5-0.75 Hz, and also medium values for the mean and various quantiles. 

In addition to looking at the slope from a linear regression on speech dynamic complexity, we also investigated the change between successive windows. Consistently with the previous target, the average dynamic complexity seems to increase the most (compared to the previous time window) when body movement has high complexity, at least according to Fourier entropy, CID CE and the number of peaks. This is consistent with the hypothesis generated when looking at the regression slope, namely that dynamic complexity of speech increases fastest when multiple measures of body movement complexity are already high. We also observed patterns where the frequency components at 0.0-0.25 Hz should be low while the components at 0.25-0.75 Hz should be medium in strength. The reliability and significance of these frequency components as indicators of increasing speech complexity could be investigated through future work.

\section{Conclusions}\label{sec:conclusions}

In conclusion, we have presented a novel approach for local exceptionality detection in time series using subgroup discovery -- as a workflow that can be applied in multiple iterations for modeling features and the target in the spirit of EMM on time series data. With these, it is possible to identify features which are strongly associated with or highly predictive of a target variable. We demonstrated the approach via a case study of analysing team interaction data, matching body movement information to a measure of the dynamics of speech. Among other things, this showcased the hypothesis-generating capabilities of our approach. Future research could consider possible refinements to our approach, for example by considering the impact of extracting features from overlapping windows at different offsets as an alternative to non-overlapping windows. In addition, techniques for modeling explicit (higher-order) relations between the respective time points are another interesting direction to consider.

For the case study, future work should further investigate hypotheses generated via the discovered subgroups. These give a suggestion as to which aspects of body movement might be predictive of changing speech dynamics. For example, the conjunction of low strength in frequency components around 0.25 Hz and below, with a pattern of body movement that lacks clear repetitions, could be predictive of choatic speech dynamics. Iterative exploratory and confirmatory research efforts could lead to new discoveries with large time series datasets.

\end{document}